# Understanding Students' Rationales for AI-Mediated Cheating: An Instructor Perspective from Online Forums Analysis

Iman Mohammadi, Sina Rismanchian, Ali Keramati
imohamm1@uci.edu, srismanc@uci.edu, a.kera@uci.edu
University of California, Irvine

Shayan Salehi, s4salehi@uwaterloo.ca, University of Waterloo

**Abstract:** The extensive adoption of AI has precipitated a crisis in academic integrity. While studies attempt to quantify cheating, less is known about the motivations instructors believe students have for using AI in prohibited ways. We analyze 2.01 million posts on r/Professors to understand how instructors perceive students' justifications for AI-mediated cheating. Our qualitative analyses reveal four primary themes: (1) Academic System Gaming, (2) Digital-Academic Literacy Gap, (3) Performance Anxiety, and (4) Normalization of Dishonesty.

## Introduction

Large Language Models (LLMs) such as ChatGPT have rapidly reshaped the educational landscape, challenging conventional notions of authorship, learning, and assessment integrity (Kazley, 2024). Within months of their release, generative AI systems became ubiquitous in higher education, adopted by both instructors seeking innovation and students navigating unprecedented opportunities for automation (Balabdaoui et al., 2024). Yet, this technological disruption has also precipitated an academic integrity crisis, as instructors grapple with detecting and interpreting new forms of AI-mediated misconduct (Yusuf, 2024; von Garrel, 2023).

Despite this growing conversation, instructors' perspectives on student motivations remain underexamined. Most existing research privileges student self-reports, which may obscure how instructors interpret and respond to the motivations they perceive (Sumakul, 2022). Yet, instructors play a crucial role in framing academic integrity, shaping institutional responses, and constructing the moral discourse surrounding AI use in classrooms (Wenger, 1998; Lave & Wenger, 1991). Understanding how instructors conceptualize students' reasons for cheating with AI provides insight not only for emerging pedagogical challenges, but also into the broader socio-technical negotiation between automation, learning, and trust (Reddy-Lund, 2025; Scardamalia, 2014).

To address this gap, we analyze 2.01 million posts (including threads and comments) from the r/Professors subreddit to examine how instructors discuss, interpret, and theorize their students' justifications for AI-mediated cheating. Through qualitative thematic analysis, we identify four overarching rationales instructors attribute to students (Braun & Clarke, 2006).

## Background

Early responses to AI in education have focused primarily on detection and policy. A significant body of work investigates how to identify AI-generated text or how to craft institutional guidelines that preserve assessment fairness (Shepard, 2000). Parallel efforts have sought to quantify the prevalence of AI cheating through surveys and controlled experiments, revealing widespread use of generative tools across disciplines. However, these quantitative efforts, while informative, leave underexplored the why, the underlying motivations and rationalizations that shape students' use of AI in prohibited ways (Sozon, 2024).

Prior qualitative research has begun to uncover students' moral reasoning around academic dishonesty more broadly. Studies on plagiarism, contract cheating, and collusion identify themes of instrumentalism, performance pressure, and systemic distrust in higher education. With the emergence of AI, these rationales are amplified and transformed: the boundary between legitimate assistance and unethical delegation becomes ambiguous, particularly when AI is framed as a "study aid" rather than a "substitute". Moreover, recent interviews and surveys show that students often misunderstand the epistemic status of AI outputs, treating them as valid sources of knowledge rather than synthetic predictions (Sun, 2025).

## Methodology

To address the mentioned gap, in this study, we used data from the r/Professors subreddit, a public online forum where university instructors discuss teaching, assessment, and classroom challenges. Reddit was chosen because it provides naturally occurring discussions that capture instructors' real-time reflections and debates about academic integrity in the age of AI.

## Data Collection

We searched posts and comments on cheating and AI use published after the release of ChatGPT in late 2022, when public conversation about AI cheating began to rise. Using Reddit's public API, we collected a total of 2.01 million posts (including threads and comments). To identify relevant discussions, we applied keyword-based searches using terms such as academic integrity, cheating, plagiarism, AI detection, ChatGPT, LLM, GPTZero, Turnitin, AI policy, and academic misconduct.

## Data Analysis

From the full dataset, 5 random sample pools, each containing 200 threads and comments combined, were selected to ensure diversity in topics and time periods. The selected posts were analyzed using qualitative thematic analysis to identify patterns in how instructors describe students' reasons for using AI unethically. Each sample pool was independently coded by two researchers, with disagreements resolved through discussion until consensus was reached. The analysis followed an inductive approach; initial open coding of posts was followed by grouping similar ideas into broader themes and subthemes through iterative review and discussion among the researchers.

## Results

As shown in Table 1, our analysis finds that instructors attribute students' AI-related cheating to a range of personal, cognitive, and cultural factors. The findings are organized into four major themes that reflect different ways instructors interpret students' motivations.

First, instructors believe many students treat education instrumentally, gaming the system to maximize rewards while minimizing effort, often exploiting assignment types or grading structures where AI use is hard to detect. Second, they observe a digital-academic literacy gap, in which students misunderstand academic conventions or overestimate AI's capabilities, using it as a "study aid" without recognizing its ethical or epistemic limits. Third, instructors perceive performance anxiety and skill deficits as key drivers, describing how students turn to AI as a coping mechanism for insecurity or under-preparedness. Finally, many instructors view dishonesty as becoming normalized, noting that AI-assisted cheating is increasingly seen by students as common, low-risk, and socially acceptable. These four themes, together, reveal how instructors make sense of AI-mediated misconduct not only as individual wrongdoing but also as a symptom of deeper tensions in educational systems and cultures.

**Table 1**

*Thematic Framework of Student Rationales for AI-Mediated Cheating, as Reported by Instructors*

| | Sub-themes | | |
|---|---|---|---|
| Academic System Gaming & Instrumentalist Motivation | 1A: Exploiting Systemic Vulnerabilities Targets low-detection, high-reward assignments. | 1B: Efficiency Optimization AI used as a time-saver, bypassing learning. | 1C: Gaming Admissions and Credentialing Uses AI to fabricate or enhance application materials. |
| Digital-Academic Literacy Gap & Misguided Tool Use | 2A: Misunderstanding Scholarly Synthesis Believes AI generation is legitimate brainstorming/synthesis. | 2B: Over-reliance on "Polishing" Tools Views AI as an advanced grammar/style checker, blurring authorship. | 2C: Misunderstanding Evidence and Academic Discourse: Presents AI-generated fabrications as legitimate evidence. |
| Performance Anxiety and Capability Deficits | 3A: Masking Inadequate Preparation Used due to a lack of foundational knowledge or skills. | 3B: Fear of Inadequate Performance Capable students use AI from anxiety that their work isn't "good enough." | |
| Normalization of Dishonesty and Diminished Consequences | 4A: The "New Normal" of Cheating AI cheating is perceived as a standard, unremarkable student behavior. | 4B: Boldness in the Face of Accusation Students persistently deny, obfuscate, or argue when confronted. | |

Our analysis reveals a complex landscape of student rationales for AI-mediated cheating from the instructors' perspective, categorized into four primary themes.